\documentclass{article}

\usepackage[preprint]{neurips_2026}

\usepackage[utf8]{inputenc} 
\usepackage[T1]{fontenc}    
\usepackage{hyperref}       
\usepackage{url}            
\usepackage{booktabs}       
\usepackage{amsfonts}       
\usepackage{nicefrac}       
\usepackage{microtype}      
\usepackage{xcolor}         
\usepackage{amsmath}
\usepackage{graphicx}
\usepackage{cleveref}
\usepackage{multirow}
\usepackage{float}
\definecolor{revgreen}{rgb}{0.0,0.0,0.0}
\newcommand{\rev}[1]{\textcolor{revgreen}{#1}}
\title{Large Language Models with At Most One Spike per Neuron}

\author{%
  Zhuoya Zhao \\
  \texttt{zoezhao0427@gmail.com} \\
  \And
  Parsa Omidi \\
  \texttt{parsa.omidi1@huawei.com}\\
  \And
  Aref Jafari \\
  \AND
  Richard Naud \\
  \texttt{rnaud@uottawa.ca} \\
}

\begin{document}

\maketitle

\begin{abstract}
  Leveraging their inherent sparse event-driven computation, spiking neural networks (SNNs) offer a promising path toward energy-efficient large language models (LLMs). Time-to-first-spike (TTFS) coding generates at most one spike per neuron within a time window, yielding extremely low firing rates. However, conventional TTFS SNNs are  restricted to specific structures, making it challenging to encode certain blocks in LLMs—such as layer normalization and matrix multiplications—using TTFS. To overcome this limitation, we introduce a reference-based strategy specifically to encode the four core LLM components: embedding layers, layer normalization, attention-related operations and dropout. We construct a fully TTFS-based SNN architecture and train it end-to-end. Experiments on modern LLMs like BERT and GPT-2 demonstrate that our approach achieves performance comparable to ANN counterparts on \rev{natural language understanding and common-sense reasoning, while a clear gap remains on language modeling perplexity}. To the best of our knowledge, this is the first work to scale a spiking LLM to 1.5 billion parameters using TTFS coding. \rev{We also report an estimate of spike-related energy; this is a spike-count proxy under an established cost model rather than a measurement on neuromorphic hardware.}
\end{abstract}

\section{Introduction}
As an alternative to conventional ANN-based computation, spiking neural networks (SNNs) have gained increasing attention because their event-driven processing relies on sparse and binary spikes, which makes them naturally compatible with energy-efficient neuromorphic hardware.
In particular, neuromorphic chips designed for spiking computation leverage in-memory computing to avoid frequent data transfer between memory and processors, thereby accelerating computation \cite{merolla2014million}. This characteristic of energy efficiency gives SNNs the potential to scale large language models (LLMs) for practical implementation.

Compared to the time-average rate code \cite{gerstner2014neuronal}, the population average rate code \cite{depasquale2023centrality} and the temporal firing patterns \cite{friedenberger2023silences}, Time-to-First-Spike (TTFS) coding encodes information using strictly at most one spike per neuron, yet it enables SNNs to maintain high performance by leveraging precise spike timing \cite{stanojevic2024high}.
Due to the sparsity and energy efficiency of TTFS coding, recent microelectronic research has demonstrated its computational advantages and developed dedicated mixed-signal vector-by-matrix multiplication (VMM) and neuron circuits that naturally align with the algorithmic principles of TTFS \cite{bavandpour2019energy,widmer2023design}, opening up broad prospects for the application of TTFS in edge intelligence.
\begin{figure}[!h]
	\centering
	\includegraphics[width=0.45\textwidth]{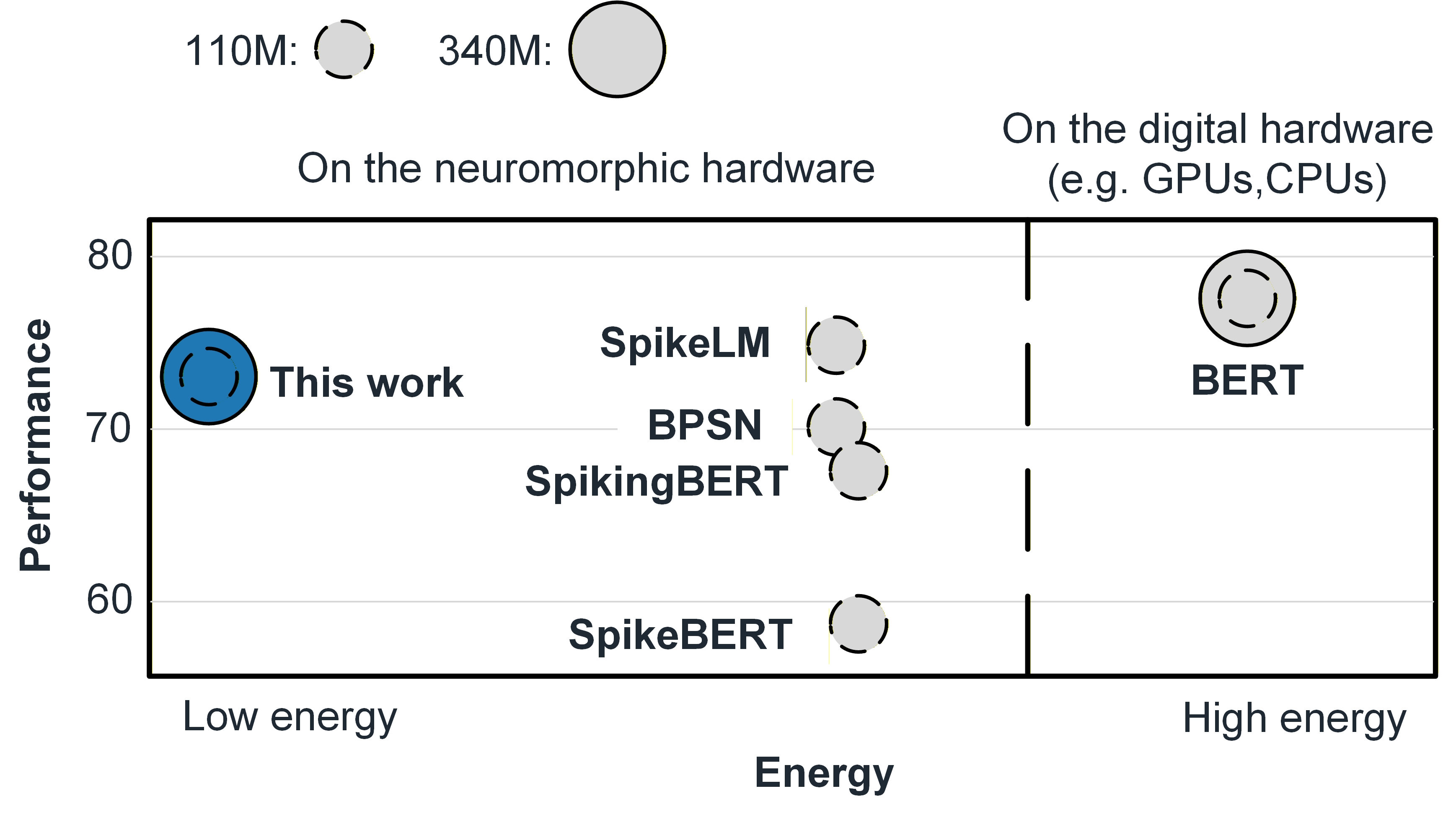}
    \caption{Comparison of performance, estimated \rev{spike-related} energy consumption (refer to Section\ref{enconan}), and parameter size for TTFS-BERT, standard BERT, and other spiking BERTs on the GLUE benchmark.}
	\label{intro}
\end{figure}
Despite its inherent sparsity, TTFS coding faces significant challenges when applied to LLMs. Prior works have demonstrated that TTFS can establish an exact mapping for a specific class of artificial neural networks, namely Multi-Layer Perceptrons (MLPs) with Rectified Linear Units (ReLUs) \cite{rueckauer2018conversion, stanojevic2023exact, stanojevic2024high}. The first spike timing is not a continuous value, since the output spike timing is delayed relative to the input, and any infinite first spike timing is usually disregarded. Consequently, the first spike timing is inherently upper-bounded, and if it exceeds this limit, it is clipped to the predefined maximum value. This constitutes the first major limitation of TTFS: the restricted range of output spike timings. Moreover, because TTFS mapping is only feasible for specific network structures, it is challenging to implement more complex components \cite{stanojevic2024high}, such as embedding layers, layer normalization (LayerNorm), and matrix multiplication, using conventional TTFS neurons.

Recent work has explored latency coding for converting ANNs into spiking transformers \cite{jiang2024spatio, zhaottfsformer}. These conversion-based approaches, may suffer from approximation errors, potentially impacting inference accuracy. To address this, we advocate for designing ANNs with inherent SNN-friendly properties—such as spike-compatible activations, which significantly enhances both accuracy and hardware efficiency. Consequently, we design a native TTFS-based LLM and train it from scratch. Our main idea is to construct an exact mapping between TTFS layers and certain components in LLMs (e.g., linear layers). For components that do not admit an exact mapping, we instead approximate them using TTFS layers (e.g., LayerNorm).
Evaluating this architecture requires rigorous consideration of training dynamics, particularly as we scale from small networks to LLMs. To validate the scalability of our approach, we conduct comprehensive experiments on LLMs across varying parameter scales.

Scaling SNNs for computational purposes ultimately targets their deployment on neuromorphic chips. Therefore, two key questions must be addressed: first, how to realize TTFS-based LLMs; and second, how to maintain competitive performance comparable to their ANN version. To the best of our knowledge, this is one of the first works to explore TTFS coding for large-scale language models. Our main contributions are summarized as follows:

(1) 
We introduce a reference-time mechanism in LLMs that enhances the expressiveness of TTFS neurons, enabling signed activations.

(2) We extend reference-based TTFS (R-TTFS) encoding to four key modules (embedding layers, LayerNorm, matrix multiplication, dropout) in LLMs, enabling TTFS-based computation throughout the entire LLM pipeline.


(3) We show that our method scales effectively to LLMs such as BERT and GPT-2, \rev{reaching ANN-level performance on natural language understanding and commonsense reasoning benchmarks}. 

All these contributions \rev{point toward} spiking LLMs \rev{whose spike count, and hence estimated spike-related energy, is substantially lower than that of rate-coded spiking baselines} (\rev{Figure} \ref{intro}).

\section{Related  works}
\subsection{Neural Coding Schemes}
There are four common coding schemes in SNNs: time-average rate code, population average rate code \cite{depasquale2023centrality}, temporal firing patterns \cite{friedenberger2023silences} and latency coding (e.g. TTFS). From the perspective of time-average rate code, the fundamental unit of information and computation is the firing rate, which experimentally measures a finite number of spikes within a limited time window \cite{gerstner2014neuronal}.
To reduce the required counting time, populations of neurons with noise can estimate the firing rate over a shorter time window. Employing population coding requires additional memory and poses challenges for scaling up. Burst coding is a neural coding scheme in which information is represented by a rapid sequence of spikes rather than by single spikes or average firing rates.
Alternatively, temporal coding identifies precise spike timing as the fundamental computational unit. TTFS uses the timing of the first spike to encode information, utilizing at most one spike, making it a very sparse coding scheme.
The main advantage of TTFS are energy efficiency \cite{davies2021advancing} driven by low firing rates. 

\subsection{Evolution and Principles of TTFS Coding}
The basic idea of TTFS is to only consider the time of the first threshold crossing \cite{gerstner2002spiking}.
Absolute and relative spike timings are two approaches for implementing TTFS coding in SNNs. In the absolute scheme, the actual spike timing is taken as the output of each layer \cite{mostafa2017supervised, goltz2021fast}, whereas in the relative scheme, the output is defined as the difference between the spike timing and a reference time  \cite{rueckauer2018conversion, stanojevic2023exact,wei2023temporal, stanojevic2024high, zhang2025toward, zhaottfsformer}. Because the relative scheme operates within a fixed time range, while the absolute spike timing accumulates across layers, the relative scheme is more stable for scaling up neural networks.

\subsection{Challenges in TTFS-based LLMs}
Conventional TTFS coding has been limited to non-negative outputs, which prevents its direct application to common LLM components like LayerNorm and matrix multiplications. Consequently, it cannot directly satisfy the signed representational requirements of these modules. One feasible approach is to use an alternative reference time in place of the upper-bound time in order to generate negative values \cite{zhaottfsformer}. To approximate the complex operations, Jiang et al. proposed a method to compute high-dimensional operations by unrolling them into several steps \cite{jiang2024spatio}. 
However, they ignored the temporal relationship between spikes and did not explain how to use spike-related variables to represent the mean and variance. Although these variables can be interpreted in terms of firing rates, the corresponding spike trains are not explicitly available. To ensure compatibility with neuromorphic hardware, it is essential that these variables be represented directly through spike-based encoding. 
Another method is to modify the differential function of TTFS neurons so that, after integration, the differentiable function can match the nonlinearities used in ANNs \cite{zhaottfsformer}. Complex differential functions are difficult to realize precisely on the neuromorphic chips because their nonlinearities rely on Resistor–Capacitor (RC) circuits, Bipolar Junction Transistors, or memory resistor characteristics, whose parameters are intrinsic and hard to adjust \cite{garg2024versatile}. Researches on SNNs need to focus not only on theoretical aspects of algorithms and architectures but also on practical implementations on neuromorphic chips. 

{Implementing matrix multiplication using TTFS is  challenging.} One approach converts spike timing multiplication into logarithmic addition \cite{zhaottfsformer}, but computing logarithms on neuromorphic hardware \cite{davies2021advancing, 10.3389/fnins.2022.795876} incurs higher energy overhead due to multi-step approximations and memory accesses.
Another method leverages membrane potentials and temporal splitting to accumulate multiplication results \cite{jiang2024spatio}, but it is prone to estimation errors. These limitations motivate the development of more accurate and energy-efficient TTFS-based matrix multiplication schemes.

\section{Preliminary}
\subsection{Time-to-first-spike model}
\begin{figure}[H]
	\centering
    \includegraphics[width=0.45\textwidth]{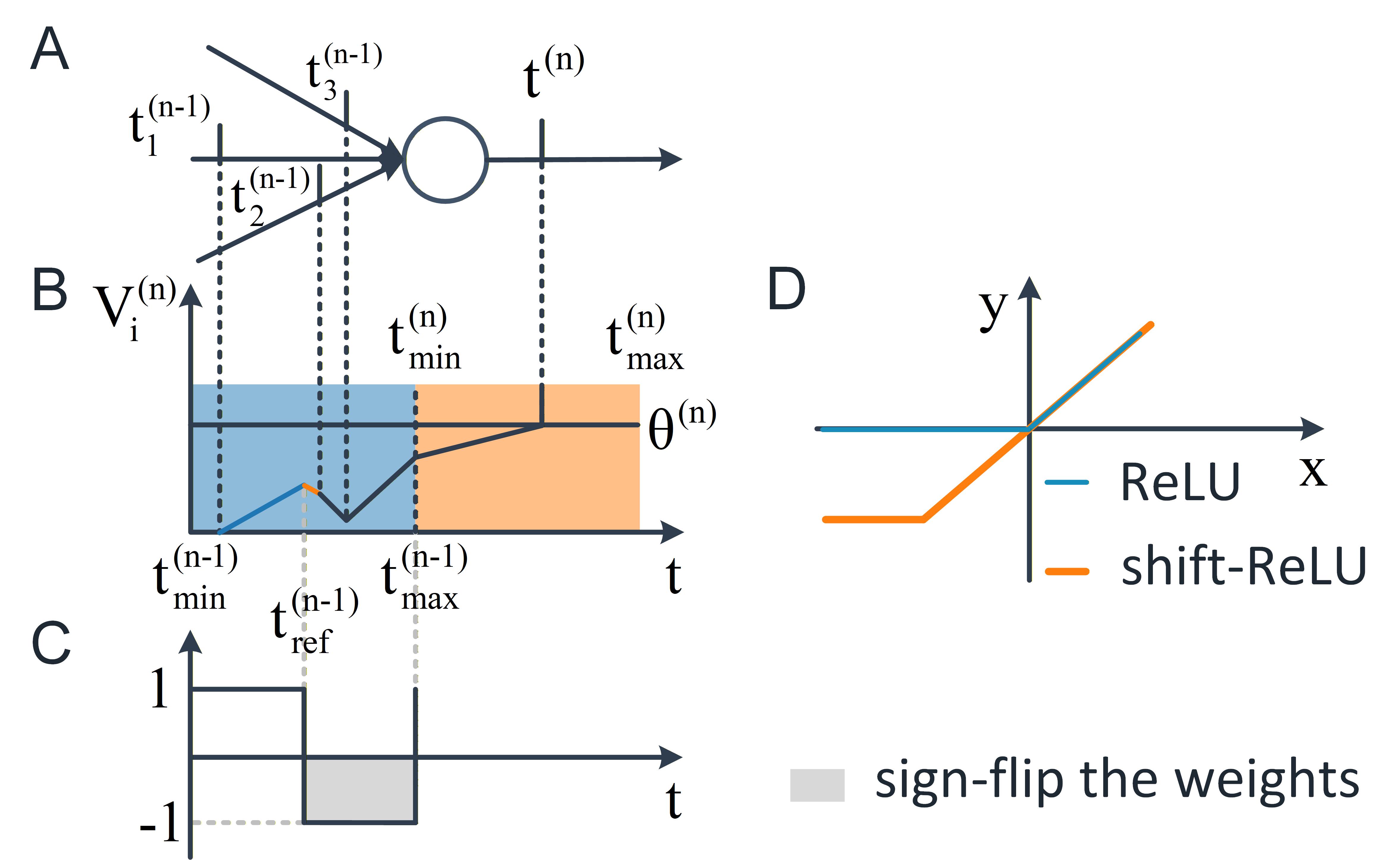}
	\caption{A. Spikes propagation. B. Integral of membrane potential $V_i^{(n)}$ over time. The blue background represents $t \in (t_{min}^{(n-1)}, t_{max}^{(n-1)})$, and the orange background represents $t \in (t_{min}^{(n)}, t_{max}^{(n)})$. The color change (from blue to orange) of the slope corresponds to the variation of the gates. C. Gating signal. D. ReLU and its variant. }
	\label{midtime}
\end{figure}

TTFS is a case of latency coding and it considers the precise latency between the beginning of a stimulus and the first spike emitted by a neuron \citet{rueckauer2018conversion, goltz2021fast, stanojevic2023exact, stanojevic2024high}. The TTFS neuron dynamics are described by a piece-wise function comprising two phases: input accumulation and threshold crossing (Equation \ref{soma}), which governs how the potential $V_i^{(n)}$ evolves over time $t$. $A_i^{(n)}$, $B_i^{(n)}$ and $\tau_{c}$ are hyperparameters and $W_{ij}^{(n)} $ is a trainable parameter. $H$ is the Heaviside step function. And $t_{min}^{(n)}$ is equal to $t_{max}^{(n-1)}$.
\begin{equation}
	\tau_c \frac{dV_i^{(n)}}{dt} =
	\begin{cases}
		A_i^{(n)} + \sum_j W_{ij}^{(n)} H\!\bigl(t - t_j^{(n-1)}\bigr),
		 \qquad \text{for } t < t_{\min}^{(n)} \\[4pt]
		B_i^{(n)},
	\qquad \text{for } t_{\min}^{(n)} \le t \le t_{\max}^{(n)}
	\end{cases}
	\label{soma}
\end{equation}
As illustrated Figure \ref{midtime}A and \ref{midtime}B, the spike timings $t_1^{(n-1)},t_2^{(n-1)},t_3^{(n-1)}$ from the previous layer ($n{-}1$) drive integrate-and-fire dynamics that generates the membrane potential $V_i^{(n)}$, shaped by a linear synaptic kernel. This dynamic process ensures that at most one spike occurs within the time window $[t_{\min}^{(n)}, t_{\max}^{(n)}]$. The membrane potential is reset when a spike occurs or $t \geq t_{max}$. 

We consider the special case with $A_i^{(n)} = 0$ and $B_i^{(n)} = 1$, commonly referred to as the \textbf{$\mathbf{B1}$-model}.
By taking the integral of the neuron dynamics, the fundamental relationship between the input $t_j^{(n-1)}$ (stimulus timing) and the output $t_i^{(n)}$ (first-spike timing) of a linear layer is derived over the interval $t_{\min}^{(n)} \le t_i^{(n)} \le t_{\text{max}}^{(n)}$, where $\theta$ denotes the firing threshold (Equation \ref{in0}), enabling an exact mapping from ReLU-based ANNs to TTFS SNNs \cite{stanojevic2024high}.
\setlength{\abovedisplayskip}{0.1in}
\setlength{\belowdisplayskip}{0.1in}
\section{Methods} \label{method}
As outlined in the Preliminary, TTFS SNNs are exactly mapped from ReLU-based ANNs. However, this formulation imposes strict constraints on the output range and model architectures, as the output timing is inherently bounded by $t_{\min}^{(n)}$ and $t_{\max}^{(n)}$, and the integration dynamics are strictly linear with ReLU.
Specifically, the architecture of LLMs extends beyond simple ReLU activations to incorporate a diverse set of components, including linear projections, LayerNorm, and attention mechanisms, all of which are essential to their overall behavior. However, conventional TTFS SNNs fail to address these complex architectural elements \cite{stanojevic2024high}.

Therefore, in this section, first, we extend the $B1$-model to a R-TTFS variant and establish it as the fundamental building block for TTFS SNNs.
We simulate the four main components—embedding layer, layer
normalization, attention mechanism, and dropout—using the R-TTFS neuron. 


\subsection{Reference-based TTFS neuron}
\subsubsection{Expanding output range via reference time}  To encode negative values via relative temporal shifts, we introduce a reference time $t_{ref}$ to TTFS (Figure \ref{midtime}B).
\begin{equation}
	\label{in0}
	t_{\text{min}}^{(n)} -  t_{i}^{(n)}  = \sum_{j=1}^N W_{ij}^{(n)} \bigl( t_{\text{max}}^{(n-1)} - t_j^{(n-1)} \bigr) -  \tau_c \theta^{(n)}_{i}
\end{equation}
To enable signed inputs and outputs, we set the reference point to the midpoint between $t_{\min}$ and $t_{\max}$, yielding Equation \ref{mid2}, where $\Delta = \frac{1}{2}(t_{\max} - t_{\min})$.
\begin{equation}
	\label{mid2}
	\begin{aligned}
		t_{\text{ref}}^{(n)} - t_{i}^{(n)}
		&= \sum_{j=1}^N W_{ij}^{(n)}
		\bigl( t_{\text{ref}}^{(n-1)} - t_j^{(n-1)} \bigr) 
		 + \sum_{j=1}^N W_{ij}^{(n)} \Delta
		+ \Delta
		- \tau_c \theta^{(n)}_{i}
	\end{aligned}
\end{equation}
\subsubsection{Exact mapping via the linear regime of shift-ReLU} To derive the mapping relationship between R-TTFS SNNs and ANNs, we introduce shift-ReLU, which corresponds to a standard ReLU shifted negatively along $y=x$ as shown in  Figure \ref{midtime}D. Because shift-ReLU allows both positive and negative inputs and outputs in its linear regime, the exact mapping between R-TTFS SNNs and linear layers of ANNs ($y=w_{ij}^{(n)}x + b_{i}^{(n)}$) still holds within this regime. 
For the given the weights $W_{ij}^{(n)}$ and the threshold $\theta^{(n)}_{i}$, the corresponding weight matrices $w_{ij}^{(n)}$ and  bias vectors $ b_{i}^{(n)}$ in the equivalent linear regime of shift-ReLU network can be expressed as follows:
\begin{equation}
	\begin{aligned}
		w_{ij}^{(n)} &\stackrel{\text{def}}{=} W_{ij}^{(n)}, \\
		b_{i}^{(n)} &\stackrel{\text{def}}{=} \sum_{j=1}^N W_{ij}^{(n)}\Delta + \Delta - \tau_c\theta^{(n)}_{i}.
	\end{aligned}
	\label{def}
\end{equation}
Essentially, this method establishes an exact mapping with a shift-ReLU ANNs in the linear regime of the shift-ReLU. 
Based on R-TTFS and its variants, we can convert each component of LLMs into TTFS SNNs. The \textbf{four} main components are the embedding layer, LayerNorm, attention mechanism, and dropout.
\subsection{Component-level Encoding of LLMs via TTFS}
\subsubsection{TTFS Embedding layers}
\begin{figure}[!h]
	\begin{center}	\includegraphics[width=0.45\textwidth]{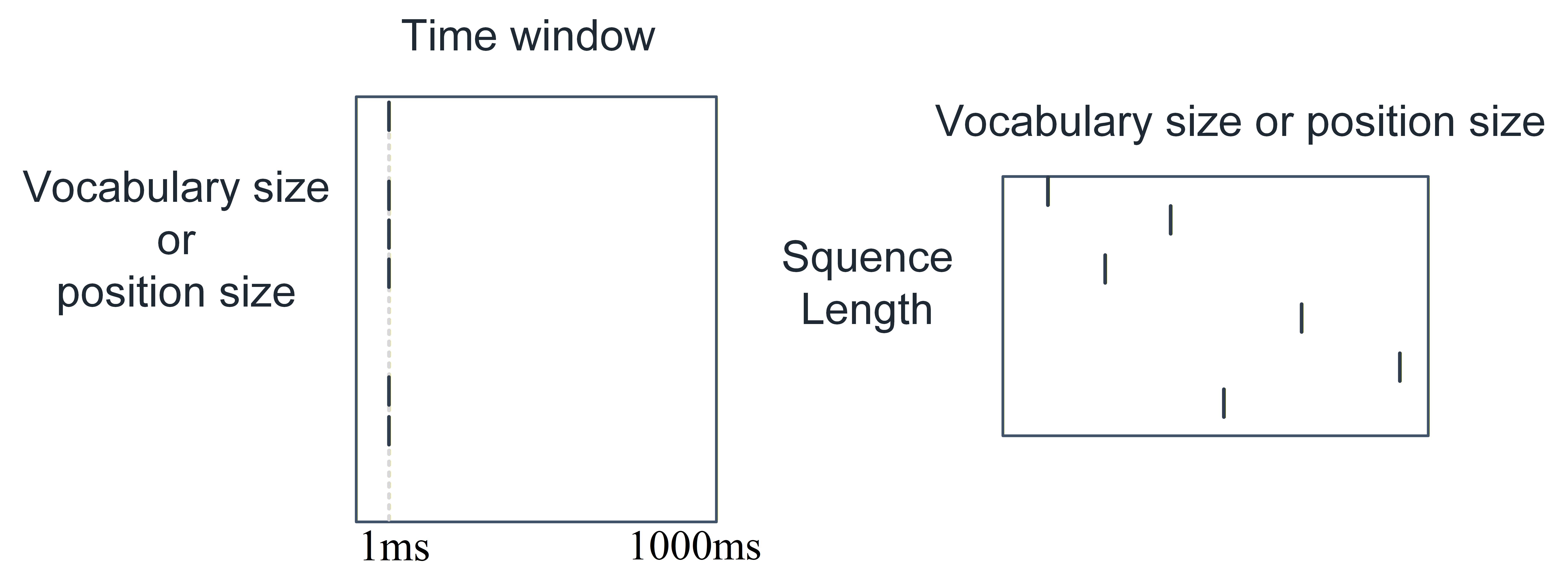}
	\caption{ Left. A set of neurons representing the vocabulary or positions are waiting to be activated by input tokens. The time window is set to $1000ms$. A spike is generated if the corresponding token ID or position exists. The spike timings are set to $1ms$. Right. Each token or position in the input sequence activates the corresponding neuron.}
	\label{embedding}
	\end{center}
\end{figure}

In contrast to conventional embedding retrieval via table lookup, the TTFS embedding layer computes representations through sparse spike-timing-based multiplication. Specifically, input token IDs or positions are one-hot encoded, where an active entry (1) corresponds to a spike emitted at a user-defined $1m s$, while an inactive entry (0) implies silence (Figure \ref{embedding}, Left). Given these input timings, the output spike timings of the embedding layer are derived according to \rev{Equation} \ref{mid2}.
\subsubsection{TTFS layer normalization}
\begin{figure}[!h]
	\centering
\includegraphics[width=0.45\textwidth]{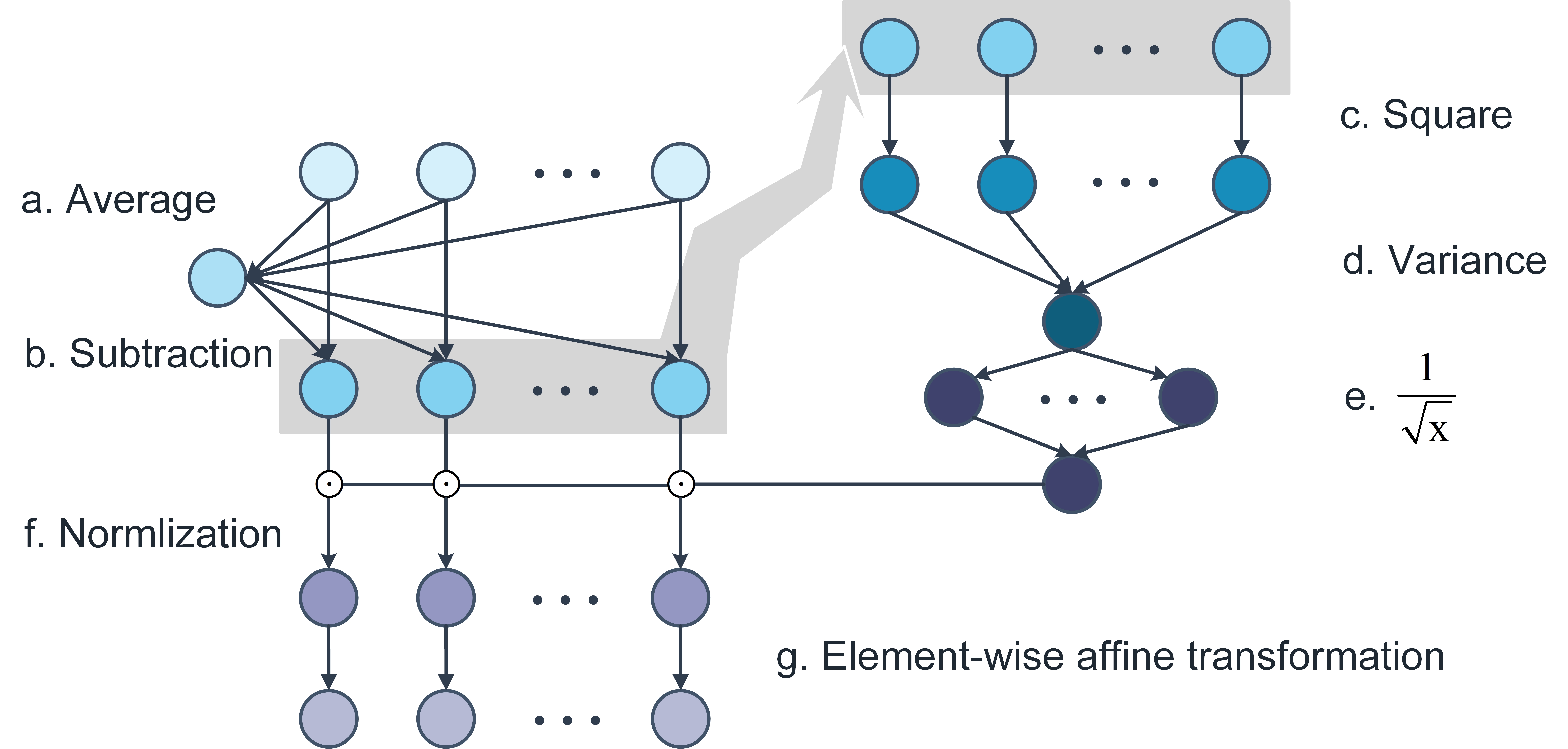}
	\caption{Integration for LayerNorm.}
	\label{layer}
\end{figure}

LayerNorm consists of a sequence of elementary operations, including mean and variance computation, normalization, and affine transformation.
In the TTFS implementation, all these operations are implemented within TTFS layers whose weights are fixed throughout LLMs training.
Accordingly, we unroll Equation \ref{layernorm} into the seven steps shown in Figure \ref{layer}.
\begin{equation}
	\mathrm{LN}(x_i) = \gamma \cdot \frac{x_i - \mu}{\sqrt{\sigma^2+\epsilon}} + \beta
	\label{layernorm}
\end{equation}
{\textbf{The mean TTFS neuron model.}}\label{meanTTFS}
To compute the mean in TTFS, the neuron must aggregate all input spikes over the entire time window and encode the resulting mean value as a spike time.
We denote this output spike time as $t_{\mu}$.
Since this aggregation requires observing all inputs within the window, the result necessarily lags behind the actual input events.
Therefore, the output spike cannot be emitted within the current window and is instead assigned to the subsequent time bin.
The dynamics of the mean neuron also follows the Equation \ref{mid2}. To enable averaging using SNNs with TTFS, it is essential to ensure that the network has fixed parameters:
\begin{equation}
	\begin{aligned}
		\frac{1}{N} \stackrel{\text{def}}{=} W_{ij}  , 
		0 \stackrel{\text{def}}{=}  \sum_{j=1}^N W_{ij}\Delta + \Delta -  \tau_{c}\theta_{i} .
	\end{aligned}
\end{equation}
{\textbf{The subtraction TTFS neuron model.}}
The subtraction TTFS neuron has only two input spike timings, one is $t_{\mu}$ spanning $(t_{\min}^{(n)}, t_{\max}^{(n)}]$, the other one is the corresponding input spike timings $t_j$ which lies in $(t_{\min}^{(n-1)}, t_{\max}^{(n-1)}]$.
The dynamics of the subtraction neuron follows :
\begin{equation}
	\tau_c \frac{dV_i^{(n+1)}}{dt} =
	\begin{cases}
		W_j H\!\bigl(t - t_j^{(n-1)}\bigr) - W_{\mu} H\!\bigl(t - t_{\mu}^{(n)}\bigr),
		\hfill t < t_{\min}^{(n+1)} \\[6pt]
		1,
		\hfill t_{\min}^{(n+1)} \le t \le t_{\max}^{(n+1)}
	\end{cases}
	\label{somasub}
\end{equation}
We encode the difference between each input and $t_{\mu}$ as a spike timing $t_{sub}$. 
When the parameters satisfy Equation  \ref{sub1} where  \( W_{j}, W_{\mu} \in \mathbb{R} \) are scalars, Equation \ref{sub} can simulate subtraction effectively.
\begin{equation}
\label{sub}
    t_{ref}^{(n+1)} - t_{sub}^{(n+1)} = W_j \!\bigl( t_{ref}^{(n-1)} - t_j^{(n-1)} \bigr) - W_{\mu} \!\bigl( t_{ref}^{(n)} - t_{\mu}^{(n)} \bigr) + (3W_j + W_{\mu} + 1) \Delta - \tau_c \theta^{(n+1)}_i
\end{equation}
\begin{equation}
	\label{sub1}
	\begin{aligned}
		1 \stackrel{\text{def}}{=} W_j,
		1 \stackrel{\text{def}}{=} W_{\mu},
		0 \stackrel{\text{def}}{=} \bigl(3W_{j} + W_{\mu} + 1\bigr)\Delta - \tau_c\theta^{(n+1)}_{i} .
	\end{aligned}
\end{equation}
{\textbf{The square TTFS neuron model.}}
Each square TTFS neuron receives input from only one presynaptic neuron. Suppose the input $t_{j}$ belongs to $ (t_{\min}^{(n-1)}, t_{\max}^{(n-1)}] $.
The dynamics of the square neuron is defined by Equation \ref{square} where $W$ is a scalar.
\begin{equation}
	\label{square}
	{\renewcommand{\arraystretch}{0.8}%
		 \frac{\tau_cdV_i^{(n)}}{dt} =
		\begin{cases}
			W\!H\!\bigl(t\!-\!t_j^{(n-1)}\bigr)\!
			\bigl(2t_j^{(n-1)}\!-t_{\min}^{(n-1)}\!-\!t_{\max}^{(n-1)}\bigr),
			\hfill  t<t_{\min}^{(n)}\\
			1, 
			\hfill  t_{\min}^{(n)} \le t \le t_{\max}^{(n)} 
	\end{cases}}
\end{equation}
Upon integrating Equation \ref{square}, the square operation can be exactly mapped to a TTFS layer (Equation \ref{sqr}). When the parameters satisfy Equation \ref{srq1}, the input $x$ and the output $y$ are equal to $ t_{\text{ref}}^{(n-1)} - t_{sub} $ and
$t^{(n)}_{ref}-t_{sqr}$, respectively.
\begin{equation}
	t_{ref}^{(n)} - t_{i}^{(n)} = W {\bigl(t_{ref}^{(n-1)}-t_{j}^{(n-1)}  \bigr)}^{2} + W\Delta^{2} +\Delta -\tau_{c} \theta 
	\label{sqr}
\end{equation}
\begin{equation}
	\label{srq1}
	\begin{aligned}
		1 \stackrel{\text{def}}{=} W,
		0 \stackrel{\text{def}}{=} W\Delta^{2} +\Delta -\tau_{c} \theta.
	\end{aligned}
\end{equation}
{\textbf{The variance TTFS neuron model.}} 
The variance is computed as the mean of all squared inputs; therefore, the process follows the same principle as the mean TTFS neuron.

{ \textbf{A TTFS-based approximation of $\frac{1}{\sqrt{x}}$ .}}

The nonlinear operations—particularly division and square root extraction—pose significant challenges for representation through neuronal dynamics' integral functions, and thus must be approximated.

\begin{figure}[!h]
	\centering
	\includegraphics[width=0.4\textwidth, trim=10bp 12bp 10bp 10bp, clip]{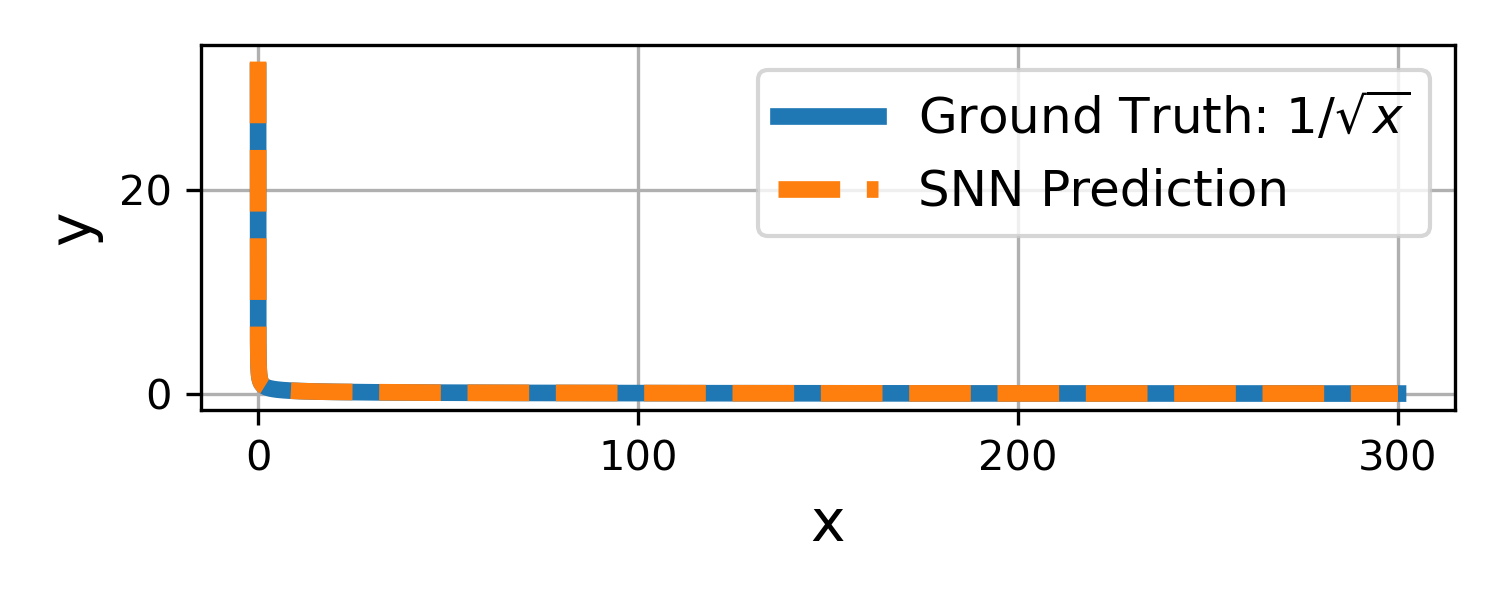}    
	\caption{Fitting $\frac{1}{\sqrt{x}}$ by using two fully connected layer.}
	\label{2f}
\end{figure}

Following the universal group operator \cite{jiang2024spatio}, we use two fully connected TTFS layers to approximate this non-linearity. 
The first layer follows the $B1$-model as defined in Equation \ref{in0}, and its outputs are encoded as spike timings. The second layer produces membrane potentials and consists of non-spiking readout neurons. The dynamics of the output neurons are modeled in \rev{Equation} \ref{v_dy}, and the membrane potential is given by Equation \ref{v1} \cite{stanojevic2024high}.
\begin{equation}
	\label{v_dy}
	\tau_c \frac{dV_i^{(n)}}{dt} = 
	\sum_j W_{ij}^{(n)} H \bigl( t - t_j^{(n-1)} \bigr)  
\end{equation}
\begin{equation}
	\label{v1}
	V_i^{(n)} = 
	\sum_j W_{ij}^{(n)} \bigl( t_{ref}^{(n-1)} - t_j^{(n-1)} \bigr) + \sum_j W_{ij}^{(n)} \Delta + D
\end{equation}
The loss is defined as the mean squared error (MSE) between the outputs of $\frac{1}{\sqrt{x}}$ and the predictions. The fitting results are illustrated in Figure \ref{2f}. During TTFS model training, we use the well-trained weights to initialize the TTFS-based approximation $\frac{1}{\sqrt{x}}$ and keep them fixed.
\textbf{The normalization TTFS neuron model.}
To simulate $\frac{x_i - \mu}{\sigma}$, the input spike timing $t_{sub}\in (t_{min}^{{n-1}}, t_{max}^{n}]$ is  multiplied by the weights ($W_{\sigma}$). When $ (W_{\sigma} + 1)\Delta - \tau_c \theta$ is equal to 0, Equation \ref{aln} can simulate $\frac{x_i - \mu}{\sigma}$.
\begin{equation}
	t_{ref}^{(n)} - t^{(n)}_{norm} = W_{\sigma}(t_{ref}^{(n-1)} - t_{sub}^{(n-1)}) + (W_{\sigma} + 1)\Delta - \tau_c \theta
	\label{aln}
\end{equation}
{\textbf{Element-wise affine transformation TTFS neuron model.}}
The operation corresponds to the Hadamard product between the spike timings $t^{(n)}_{norm}$ and the scaling parameter $\gamma$, and then add $\beta$, which means there is only one input spike in this TTFS neuron. When $W_{\sigma}$ is equal to $\gamma$ and $(W_{\sigma} + 1)\Delta - \tau_c \theta$ is equal to $\beta$, Equation \ref{aLN1} be mapped to the standard LayerNorm formulation.
\begin{equation}
	t_{ref}^{(n)} - t^{(n)}_{LN} = W\odot(t_{ref}^{(n-1)} - t^{(n-1)}_{norm}) + (W + 1)\Delta - \tau_c \theta
	\label{aLN1}
\end{equation}
\subsubsection{TTFS Attention Block}
\textbf{Attention score computation.}
$Q$ and $K$ are represented by membrane potential $V_{Q}^{(n)}$ and spike timing $t_{K}^{(n)}$, respectively.
\begin{equation}
	\label{vq1}
	V_{Q}^{(n)} = 
	\sum_j W_{qj}^{(n)} \bigl( t_{ref}^{(n-1)} - t_j^{(n-1)} \bigr) + \sum_j W_{qj}^{(n)} \Delta + D_q
\end{equation}
\begin{equation}
	\label{vk1}
	t_{\text{ref}}^{(n)} - t_{K}^{(n)} = \sum_{j} W_{kj}^{(n)} \bigl(t_{\text{ref}}^{(n-1)} - t_j^{(n-1)} \bigr) 
\end{equation}
By treating $V_{Q}^{(n)}$ as the weights, the equation \textcolor{black}{$V_{QK}^{(n)} = t_{K}^{(n)}V_{Q}^{(n)}$ } can be reformulated as:
\begin{equation}
	\label{qk}
	\tau_c \frac{dV_{QK}^{(n)}}{dt} = 
\sum_j V_{Q}^{(n)} H \bigl( t - t_{K}^{(n)} \bigr)  
\end{equation}
Matrix multiplication on neuromorphic hardware is replaced by the accumulation of spike inputs ($K$) weighted by linear synaptic kernels ($Q$). 

\textbf{Softmax.} Given that the membrane potential encodes the $QK$ product, the softmax operation can be performed directly.

\textbf{Attention aggregation.}
Similarly, $V$ can be expressed as Equation \ref{v2} and attention aggregation can be treated as Equation \ref{att}.
\begin{equation}
	t_{\text{ref}}^{(n)} - t_{V}^{(n)} = \sum_{j=1}^N W_{vj}^{(n)} \bigl(t_{\text{ref}}^{(n-1)} - t_j^{(n-1)} \bigr) 
	\label{v2}
\end{equation}
\begin{equation}
	t_{ref}^{(n+1)} - t^{(n+1)}_{attn} = V_{QK}^{(n)}(t_{ref}^{(n)} - t_{V}^{(n)}) + (V_{QK}^{(n)} + 1)\Delta - \tau_c \theta
	\label{att}
\end{equation}
\subsubsection{The TTFS-based Dropout}
The TTFS-based dropout follows the same principle as \rev{Equation} \ref{mid2}, with the key difference being that it randomly drops certain neurons according to a specified dropout rate.
\section{Experiments}
\subsection{Experimental Setup}
We validate the effectiveness of our TTFS method by applying it to two representative architectures. For the encoder-based setting, we pre-train TTFS-BERT-base (110M parameters) and TTFS-BERT-Large (340M) \cite{devlin2019bert} on Wikipedia and BookCorpus \cite{zhu2015aligning} to do next sentence prediction (NSP) and masked language model (MLM) \cite{devlin2019bert} tasks. These models are subsequently fine-tuned on the GLUE benchmark \cite{wang2018glue} and benchmarked against BERT \cite{devlin2019bert}, SpikeBERT \cite{lv2023spikebert}, BPSN \cite{su2024snn}, SpikingBERT \cite{bal2024spikingbert}, and SpikeLM \cite{xing2024spikellm}. Additionally, for the decoder-based setting, we pre-train TTFS-GPT-2 Small (117M) and TTFS-GPT-2 XL (1.5B) \cite{radford2019language} using the FineWeb-Edu dataset \cite{penedo2024fineweb}. We assess their generation capabilities by reporting perplexity (PPL) and accuracy metrics via the lm-evaluation-harness \cite{gao2021framework}. Due to the scarcity of full-scale Spiking GPT-2 models, we benchmark our approach against vanilla GPT-2 as the standard baseline, while simultaneously extending our comparison to recent RWKV-based spiking decoders, such as NSLLM \cite{xu2026neuromorphic} and SpikeGPT \cite{zhu2023spikegpt}. Detailed experimental settings are provided in \rev{Appendix} \ref{B}.
\begin{table*}[t]
	\caption{Performance comparison on the GLUE benchmark.
		F1 scores are reported for QQP and MRPC, Spearman correlations are reported for STS-B, and accuracy scores are reported for the other tasks.}
	\label{tab:glue_official}
	\setlength{\tabcolsep}{0pt}
	\begin{center}
		\begin{small}
			\begin{tabular*}{0.98\textwidth}{l@{\hspace{10pt}}|@{\hspace{8pt}}c@{\hspace{8pt}}|@{\hspace{0pt}}@{\extracolsep{\fill}}cccccccc}
				\toprule
				\textbf{Model} & \textbf{Size} & \textbf{MNLI-(m/mm)} & \textbf{QQP} & \textbf{QNLI} & \textbf{SST-2} & \textbf{STS-B} & \textbf{MRPC} & \textbf{RTE} & \textbf{Avg.} \\
				\midrule
				BERT & \multirow{6}{*}{\textbf{Base}} 
				& 84.0/79.6 & 87.4 & 86.2 & 92.5 & 88.8 & 85.2 & 68.6 & 84.0 \\
				SpikeBERT  & 
				& 71.4/71.0 & 68.2 & 66.4 & 85.4 & 18.7 & 82.0 & 57.5 & 65.1\\
				SpikingBERT  & 
				& 78.1/- & 86.8 & 85.2 & 88.2 & 81.9 & 79.2 & 66.1 & - \\
				SpikeLM  & 
				& 77.1/77.2 & 83.9 & 85.3 & 87.0 & 84.9 & 85.7 & 69.0 & 81.3 \\
				BPSN  & 
				& 74.1/- & 83.9 & 83.9 & 88.0 & 83.9 & 85.3 & 64.6 & - \\			
				
				\textbf{TTFS-BERT} & 
				& 77.1/77.6 & 87.0 & 85.2 & 86.3 & 82.0 & 81.6 & 66.8 & 80.5 \\
				
				\midrule
				
				BERT & \multirow{2}{*}{\textbf{Large}}
				& 82.1 /82.6  & 87.3  & 89.8 & 90.9  & 86.3 & 89.7 & 73.4 & 85.3\\                          
				\textbf{TTFS-BERT} & 
				& 81.6/80.9 & 86.3  & 85.6 & 90.1 & 82.6 & 84.0 & 67.4 & 82.3  \\	\bottomrule
			\end{tabular*}
		\end{small}
	\end{center}
\end{table*}

\subsection{Experimental Results}
\textbf{TTFS-BERT.} In Table \ref{tab:glue_official}, we report the GLUE classification performance of models with different sizes. On the Base-sized model, TTFS-BERT achieves an average score of 80.5, significantly outperforming early SNN conversion methods like SpikeBERT (65.1) by a large margin. It also performs comparably to the leading SNN method, SpikeLM (81.3), while surpassing it on specific tasks such as QQP (87.0 vs. 83.9). We further validate the scalability of our proposed method on the large architecture. TTFS-BERT (Large) attains an average score of 82.3, \rev{leaving a gap of 3.0 points with respect to the full-precision BERT-Large (85.3)}.
\begin{table*}[t]
	\caption{Performance comparison of conventional and TTFS-based GPT2 models on language modeling and common-sense reasoning benchmarks. All evaluations are performed using lm-evaluation-harness.}
	\label{tab:gpt2_lm_eval}
	\addtolength{\tabcolsep}{-3.5pt}
	\begin{center}
		\begin{small}
			\begin{tabular}{l|c|cc|ccccccccc}
				\toprule
				\textbf{Model} & \textbf{Size} & \textbf{Wiki.} & \textbf{LMB.} & \textbf{LMB.} & \textbf{PIQA} & \textbf{Hella.} & \textbf{Wino.} & \textbf{ARC-e} & \textbf{ARC-c} & \textbf{SIQA} & \textbf{BoolQ} & \textbf{Avg.} \\
				& & ppl $\downarrow$ & ppl $\downarrow$ & acc $\uparrow$ & acc $\uparrow$ & acc\_n $\uparrow$ & acc $\uparrow$ & acc $\uparrow$ & acc\_n $\uparrow$ & acc $\uparrow$ & acc $\uparrow$ & \\
				\midrule	
				GPT-2 & \multirow{4}{*}{\textbf{Small}} 
				& 37.4 & 40.1 & 32.6 & 62.5 & 31.1 & 51.6 & 39.5 & 22.7 & 36.6 & 48.7 & 40.7 \\
                    NSLLM & 
				& - & - & - & 59.3 & - & 51.0 & 39.9 & 23.3 & - & - & - \\
				NSLLM* & 
				& - & - & - & 60.5 & - & 50.8 & 38.3 & 22.8 & - & - & - \\
				\textbf{TTFS-GPT-2} & 
				& 40.4 & 92.6 & 25.4 & 63.9 & 34.3 & 50.6 & 47.0 & 25.7 & 37.9 & 55.1 & 42.5 \\
				
				\midrule
				GPT-2 & \multirow{4}{*}{\textbf{XL}} 
				& 20.4 & 10.6 & 51.2 & 70.5 & 50.9 & 58.3 & 51.1 & 28.5 & 40.3 & 61.8 & 51.6 \\
				NSLLM & & - & - & - & 66.6 & - & 50.8 & 45.3 &  26.8 & - & - & - \\    
				NSLLM* & & - & - & - & 63.7 & - & 50.2 & 42.1 & 23.6 & - & - & - \\	                
				\textbf{TTFS-GPT-2} & & 26.6 & 23.8 & 39.4 & 69.4 & 44.5 & 51.1 & 57.7 & 31.1 & 40.6 & 62.0  & 49.5 \\			
				\bottomrule
			\end{tabular}
		\end{small}
	\end{center}
\end{table*}
\textbf{TTFS-GPT2.} Table \ref{tab:gpt2_lm_eval} present zero-shot accuracy on commonsense reasoning benchmarks and PPL results on language modeling tasks.  Specifically, TTFS-GPT-2 (Small) outperforms the ANN baseline on common-sense reasoning benchmarks (Avg. 42.5 vs. 40.7), while the XL model maintains comparable performance (\rev{49.5} vs. 51.6). \rev{On language modeling, by contrast, the gap is substantial and we do not claim parity: WikiText perplexity degrades from 20.4 to 26.6 on XL, and LAMBADA degrades from 10.6 to 23.8 in perplexity and from 51.2 to 39.4 in accuracy. Discretising spike timings therefore does more than perturb the confidence of the output distribution: because LAMBADA accuracy is itself a rank-based metric, its drop shows that the ranking of logits is altered on tasks that hinge on a single long-range dependency, where timing-quantisation error accumulates over the whole context before the decisive token is predicted. Commonsense reasoning benchmarks, whose candidate completions are short and well separated in likelihood, are far less sensitive to this effect, which is why accuracy there is preserved while perplexity is not. Closing this language-modeling gap is, in our view, the main open problem for TTFS-based generative models. In comparison with other spiking neural networks, we aligned our benchmark with SpikeGPT, which solely evaluates perplexity on WikiText-103. Under this setting, TTFS-GPT-2 XL reaches 26.6, outperforming the 39.8 perplexity of SpikeGPT (216M parameters), while TTFS-GPT-2 Small reaches 40.4, i.e. on par with SpikeGPT using roughly half the parameters.}

\section{Discussion}

\subsection{Energy Consumption Analysis} \label{enconan}
It is oversimplistic to expect a definitive conclusion regarding the energy consumption of SNNs, as it is highly dependent on the underlying hardware. While SNNs are energy efficient on neuromorphic hardware, they often incur higher latency and energy costs during training or inference on standard GPUs.
Furthermore, it is imprudent to compare the energy consumption of SNNs and ANNs based solely on the distinction between multiply-and-accumulate (MAC) and accumulate (AC) operations \cite{xing2024spikelm,bal2024spikingbert,hwang2024spikedattention,zhaottfsformer}, as this perspective neglects the time and energy costs associated with implementing the internal dynamics of spiking neurons \cite{davies2021advancing}.
\begin{equation}
	\sum_{i',n} (T_r + 0.5(\vartheta_{i'}^{(n)})^2 C) + 0.5 \sum_{i'',n} (V_{i''}^{(n)}(t_{\max}^{(n)}))^2 C
	\label{eq:energy_cost} 
\end{equation}
\rev{We did not run our models on physical neuromorphic hardware, so the analysis below quantifies only the spike-related component of energy under an established cost model, and should be read as a proxy rather than as a measurement.}
We used \rev{Equation} \ref{eq:energy_cost} to estimate the dominant factor of energy consumption of neuromorphic hardware, which contains the transmission cost $T_r$ per spike and the charging cost determined by the capacitor $C$ \cite{stanojevic2024high}. A neuron incurs significantly higher energy costs when it spikes compared to when it remains silent ($T_r \gg 0.5(\vartheta_{i'}^{(n)})^2 C > 0.5 \sum_{i'',n} (V_{i''}^{(n)}(t_{\max}^{(n)}))^2 C$ ) \cite{stanojevic2024high}. This observation suggests that\rev{, under this cost model,} TTFS with at most one spike \rev{incurs a lower spike-related cost} than rate coding methods. 
 To further quantify this effect, we estimate the \rev{spike-related} energy consumption of various models using $N_s \cdot T_r$ as the primary metric, where $N_s$ denotes the total spike count. \rev{This proxy deliberately excludes weight storage and off-core memory traffic, inter-core communication, and the cost of the timing circuitry needed to resolve $\delta t$. On digital asynchronous hardware these terms can dominate and do not scale with spike count, so Table \ref{tab:en} compares spike activity between spiking models and does not establish a system-level energy advantage over ANNs. Validating these estimates on a physical chip remains future work.}
 \setlength{\abovedisplayskip}{0.1in}
\setlength{\belowdisplayskip}{0.1in}
\begin{table}[H]
	\centering
	\caption{\rev{Estimated} average \rev{spike-related} energy cost per neuron per inference. \rev{Baseline entries are obtained as (firing rate)$\times$(number of time steps) from the settings reported in the respective papers: SpikeLM $0.3\times4$ and SpikingBERT $0.2\times16$. The TTFS-BERT entry is derived from Table \ref{tab:spike_count} in Appendix \ref{B}.}}	
	\begin{small}
		\begin{tabular}{c|ccc}
			\toprule
			& \rev{SpikeLM}& SpikingBERT& \textbf{TTFS-BERT} \\ \hline
			\rev{Spike-related cost} &  $1.2T_r$ & $3.2T_r$ & \rev{$0.80\,T_r$} \\ 
			\bottomrule
		\end{tabular}
	\end{small}
	\label{tab:en}
\end{table}

\section{Conclusion}
In this work, we propose a R-TTFS coding scheme tailored for LLMs, formulating TTFS layers to effectively represent or approximate the embedding layer, LayerNorm, attention mechanism, and dropout. By constraining neurons to emit at most one spike, our approach inherently guarantees high sparsity. \rev{Under a spike-count proxy, this translates into a lower spike-related cost than rate-coded spiking baselines, while the model stays competitive with its ANN counterpart on natural language understanding and commonsense reasoning. Two limitations remain open: a clear language-modeling gap, most visible on LAMBADA, and the absence of measurements on physical neuromorphic hardware, which is required before any system-level energy claim can be made.} Overall, our work opens up a promising avenue \rev{toward low-energy} inference in LLMs, bridging the gap between deep learning and neuromorphic computing.

\bibliographystyle{plainnat}
\bibliography{reference}

\appendix
\section{Technical appendices and supplementary material}

\subsubsection{Theoretical implementation on neuromorphic chips} There is a slight difference between reference-based and conventional TTFS SNNs implemented on neuromorphic hardware. In contrast to the first branch of \cref{soma}, the reference-based neuron requires a gating signal to modulate the sign of integration as shown in \cref{midtime}C. This ensures that negative inputs ($t_{ref}-t_{j} \le 0$) can be represented correctly as integration proceeds chronologically. 
\begin{equation}
	\label{in3}
	\begin{cases} 
		\int_{t_j}^{t_{\text{ref}}^{(n-1)}} W^{(n)}_{ij} H(t - t_j^{(n-1)}) \, dt, & t_{\text{min}}^{(n-1)} \leq t < t_{\text{ref}}^{(n-1)} \\ 
		-\int_{t_{\text{ref}}^{(n-1)}}^{t_j} W^{(n)}_{ij} H(t - t_j^{(n-1)}) \, dt, & t_{\text{ref}}^{(n-1)} \leq t \leq t_{\text{max}}^{(n-1)} 
	\end{cases}
\end{equation}

\section{Implementation details} \label{B}
All pre-training experiments were conducted on a cluster of eight NVIDIA H100 GPUs (80GB memory each).

\subsection{TTFS-BERT} We train BERT for 1,000,000 update steps using the AdamW optimizer with a peak learning rate of $1\times 10^{-4}$. We employ a cosine learning-rate schedule with a 2\% warm-up ratio, after which the learning rate is annealed to the end of training. For BERT models, we use the standard WordPiece tokenizer (vocabulary size 30,522). For pretraining, we set the maximum input sequence length to 512 tokens and the batch size is 512.

\begin{table}[htbp]
\centering
\caption{Average Spike Count across Transformer Blocks per Time Window}
\label{tab:spike_count}
\renewcommand{\arraystretch}{1.1}
\footnotesize
\begin{tabular}{llcc}
\toprule
\textbf{Transformer Block} & \textbf{Sub-Block} & \textbf{Num. Neurons} & \textbf{Avg. Spike Count per Neuron per Time Window} \\
\midrule
TTFS\_LayerNorm  & mean       & 1   & 1    \\
                 & subtraction& $d$ & 1    \\
                 & square     & $d$ & 1    \\
                 & variance      & 1   & 1    \\
                 & $\frac{1}{\sqrt{x}}$  & 512 & 0.25 \\
                 & norm       & $d$ & 1    \\
                 & affine     & $d$ & 1    \\
\midrule
TTFS\_Attention  & $t_q$      & $d$ & 1    \\
                 & $t_k$      & $d$ & 1    \\
                 & $t_v$      & $d$ & 1    \\
                 & $t_{qk}$   & $N$ & 1    \\
                 & $t_{qkv}$  & $d$ & 1    \\
\midrule
TTFS\_linear     &            & $d$ & 1    \\
TTFS\_dropout    &            & $d$ & 1    \\
TTFS\_res        &            & $d$ & 1    \\
TTFS\_ReLU &            & $4d$& 0.07 \\
TTFS\_linear     &            & $d$ & 1    \\
TTFS\_dropout    &            & $d$ & 1    \\
TTFS\_res        &            & $d$ & 1    \\
\midrule
TTFS\_LayerNorm  & \multicolumn{2}{l}{\textit{(Same as initial TTFS\_LayerNorm)}} & 1 \\
\bottomrule
\end{tabular}
\end{table}

\rev{\textbf{Derivation of the spike-count estimate.} Summing the third column of Table \ref{tab:spike_count}, one TTFS transformer block contains $22d+N+1028$ neurons; the constant term is $2\times(512+2)$ because each of the \emph{two} LayerNorm sub-blocks contains a $512$-neuron $1/\sqrt{x}$ approximator together with its mean and variance neurons. Weighting each row by its average spike count gives $18.28d+N+260$ spikes per block, where the non-integer contributions come from the $1/\sqrt{x}$ block ($512\times0.25$ spikes per LayerNorm) and from TTFS\_ReLU ($4d\times0.07$); every remaining sub-block emits exactly one spike per neuron by construction. For TTFS-BERT-Base ($d=768$, $N=512$) this gives $14{,}811$ spikes over $18{,}436$ neurons, i.e. $\approx0.80$ spikes per neuron per inference, which is the $0.80\,T_r$ entry of Table \ref{tab:en}. Two remarks are worth making. First, the expanded TTFS LayerNorm does not increase the total spike count relative to a rate-coded block: a rate-coded block of $\sim16d+N$ neurons firing at $1.2$ spikes/neuron emits $\approx19.2d+1.2N$ spikes, which exceeds $18.28d+N+260$ for the hidden sizes used here and grows faster in both $d$ and $N$. Second, the $1/\sqrt{x}$ approximator contributes a constant $512$ neurons per LayerNorm independently of $d$, so its relative cost vanishes as models are scaled up.}

\textbf{Evaluation Metrics.} We report specific metrics for each dataset. Accuracy measures the simple proportion of correct predictions. For tasks sensitive to class imbalance (QQP, MRPC), we utilize the F1 score, defined as the harmonic mean of precision and recall. Finally, for semantic similarity regression (STS-B), we employ Spearman correlation to assess the strength of the monotonic rank relationship between the model's scores and human judgments.

\subsection{TTFS-GPT-2} We train the model for 1,000,000 update steps with a peak learning rate of $1\times 10^{-4}$ and a batch size of 512. The learning rate follows a linear decay schedule with a 2\% warm-up ratio.
For GPT-2 models, we use the GPT-2 byte-level BPE tokenizer (vocabulary size 50,257). For GPT pretraining, we use a context window (block size) of 512 tokens.

\textbf{Evaluation Metrics.} Perplexity (PPL) and Accuracy (Acc) serve as complementary metrics to evaluate distinct dimensions of model performance. Perplexity acts as an intrinsic indicator of generative fluency, measuring the model's uncertainty in predicting the next token; lower values reflect a more robust grasp of the underlying language distribution. In contrast, Accuracy is employed for downstream tasks (such as multiple-choice questions) to assess reasoning and knowledge application. It typically quantifies the model's ability to correctly rank candidate answers based on conditional likelihoods, ensuring that the model assigns the highest probability to the correct completion rather than merely generating text.
\section{Detailed Results} \label{C}
\begin{figure}[!htbp]
	\centering
	\includegraphics[width=\textwidth]{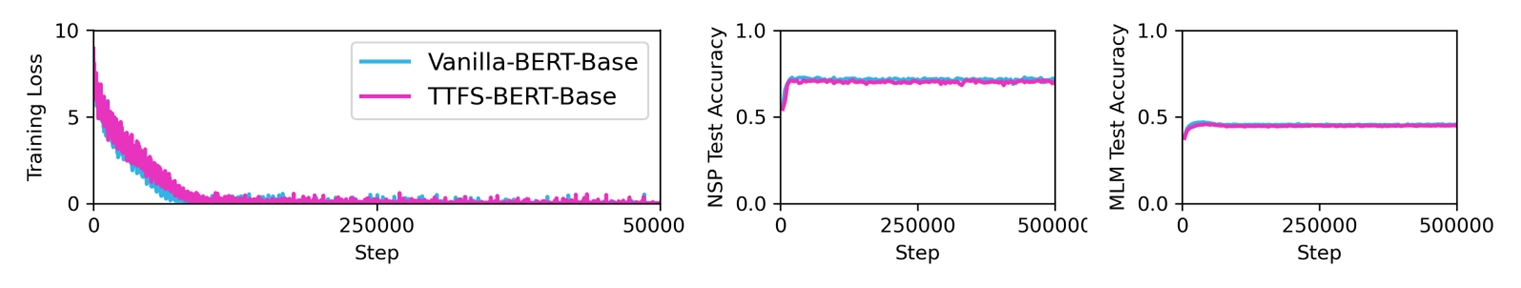}
	\caption{Comparsion between Vanilla-BERT-Base and TTFS-BERT-Base.}
	\label{bert_base_re}
\end{figure}

\begin{figure}[!htbp]
	\centering
	\includegraphics[width=0.4\textwidth]{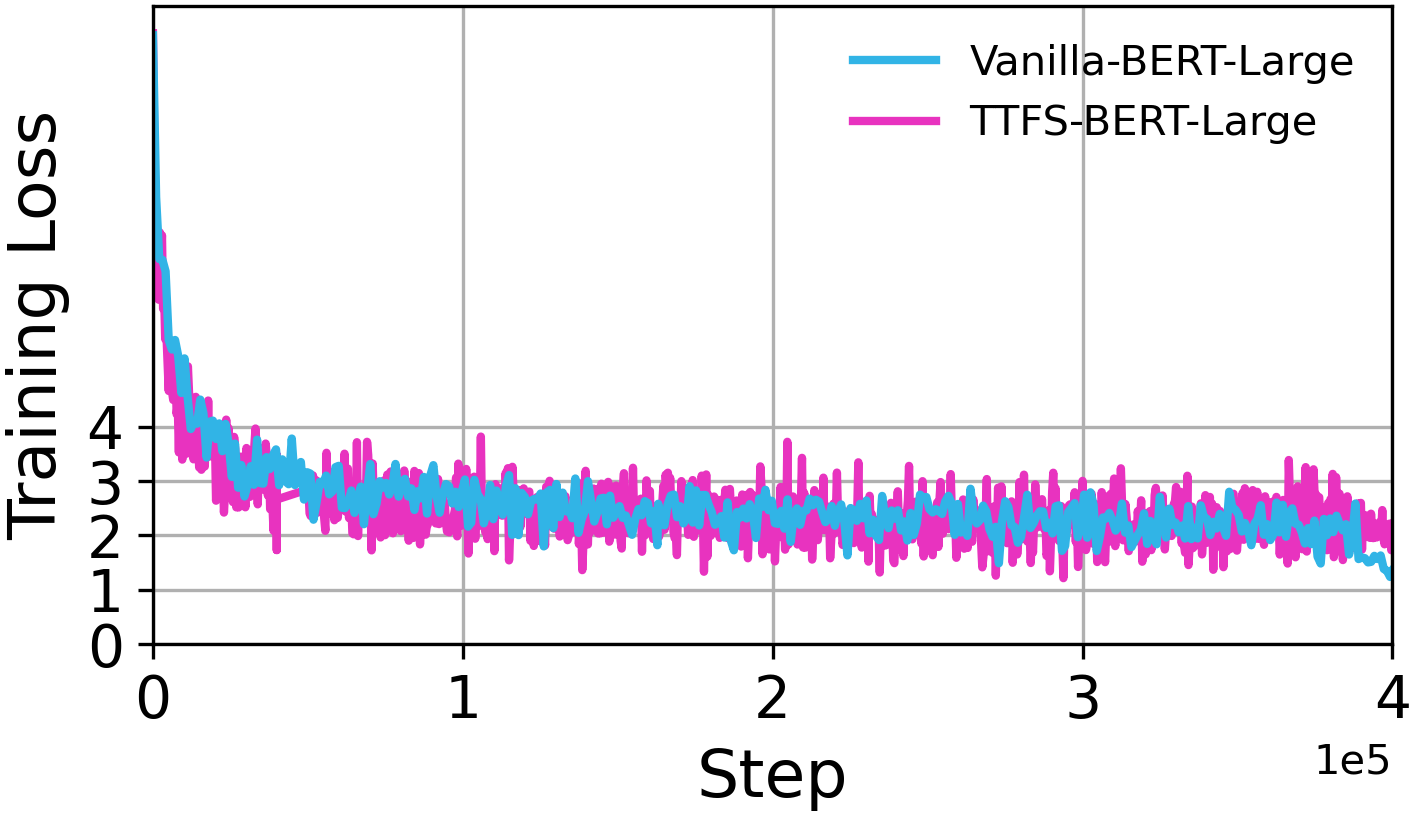}
	\caption{Comparsion between Vanilla-BERT-Large and TTFS-BERT-Large.}
	\label{1large}
\end{figure}

\cref{bert_base_re} (left) and \cref{1large} show the training loss curves for BERT-base and BERT-large, respectively.
The accuracies of the NSP and MLM tasks are shown in Fig. \ref{bert_base_re} (middle and right). Compared to BERT, the TTFS model exhibits almost no performance degradation.

\subsection{Impact of temporal configurations on representation and sparsity}

One of the advantages of TTFS coding is sparsity. In contrast to other SNN-based language models that rely on multiple spikes per unit \cite{lv2023spikebert,bal2024spikingbert,zhu2023spikegpt, xing2024spikelm, shen2025spikingssms}, our TTFS-LLMs use at most one spike per unit \rev{while remaining competitive on natural language understanding and commonsense reasoning}.
\begin{figure}[!htbp]
	\centering
	\includegraphics[width=0.4\textwidth]{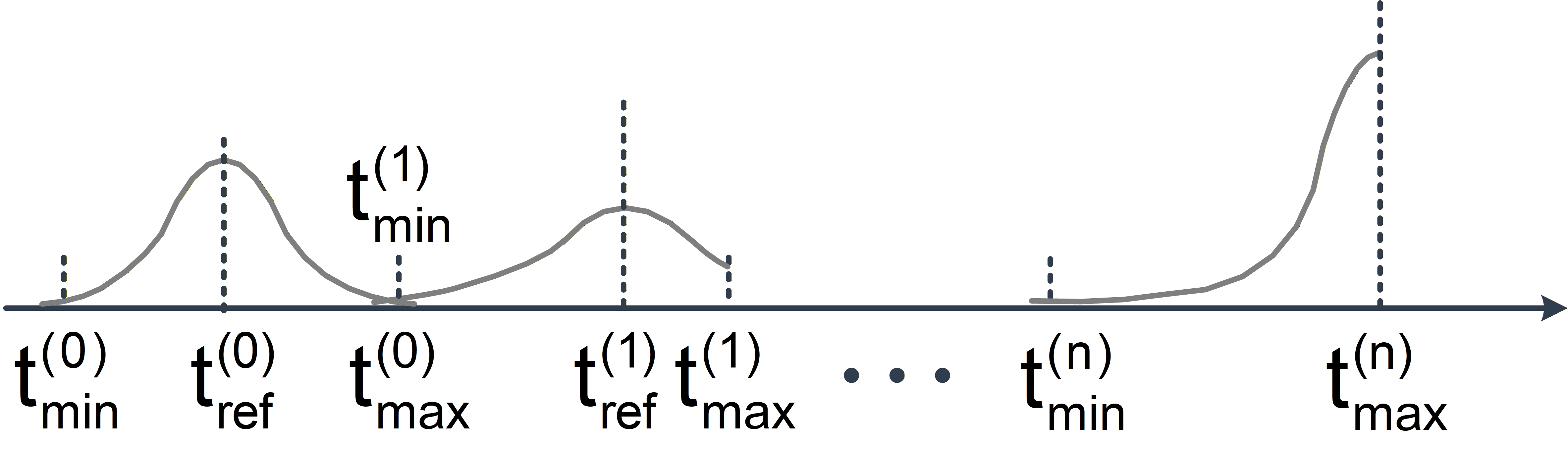}
	\caption{Schematic illustration  of the distribution of first spike timings.}
	\label{sparse}
\end{figure}

\subsubsection{R-TTFS neuron VS B1-model} Figure \ref{sparse} illustrates schematically the distribution of first-spike timings across layers. While the schematics for Layer 0 illustrate the timing distributions of symmetric R-TTFS neurons (which can be extended to an asymmetric setting in Layer 1), the schematic for Layer $n$ depicts the distribution for $B1$-model neurons, which are restricted to representing outputs with a single sign.

\subsubsection{Size of time window}
One may argue that TTFS trades off sparsity for time, such that a neuromorphic TTFS device would be too slow. The TTFS time window  ($T=t_{max}-t_{min}$) is user-defined, but must occur sequentially across layers (as shown in Figure \ref{sparse}) such that the total duration time duration is $T*M$ where $M$ is the number of layers. The considerations in choosing the time window are, like in rate-coded SNNs, a trade-off between expressibility and total latency. The expressibility of a spiking unit can be measured by the number of distinct times that can be reliably distinguished, which relates to the bit precision of the floating point number represented by the unit. For TTFS, this is the per-unit duration divided by the smallest accessible spike-timing precision $\delta t: N_{states} = T/\delta t$. For a rate-coding unit, the number of states corresponds to the number of time bins which ultimately define the minimum and maximum number of spikes, and thus again $N_{states} = T/\delta t$. Thus TTFS does not bring a longer latency per unit.

The fact that TTFS is restricted to sequentially go through each layer one at a time can, however, give rise to longer latencies than rate-coding, which does not have this fundamental constraing. It should be noted that, while rate-coding does not require layer activity to be computed sequentially, letting the rate-coding units follow absolute time is not easy to implement. Most efforts that use SNNs in the context of ANNs utilize a layer-specific time, such that there would be no advantage on the total latency \cite{zhu2023spikegpt,yao2023attention,zhou2023spikformer,shen2025spikingssms}.  Some efforts have specifically attempted to have a network follow absolute time but scaling up to LLMs remains to be proven \cite{stuck2025burst}. 

Another related consideration is that some inputs may fall into the saturated region of the TTFS mapping, producing nearly fixed spike timings and thus degrading expressivity. However, in practice, this issue hardly happens because weight initialization \cite{glorot2010understanding, he2015delving}, normalization \cite{ioffe2015batch,ba2016layer} and regularization \cite{krogh1991simple,srivastava2014dropout} help keep the outputs of each hidden layer within a reasonable range. Among them, normalization explicitly centers the outputs around zero, while initialization and regularization mainly control the scale. As a result, it becomes straightforward to adjust $t_\text{max}$ such that the output of the TTFS neurons aligns with the output of the corresponding layer in the ANNs.

\subsubsection{Firing rate}  
\begin{figure}[!h]
	\centering
	\includegraphics[width=0.48\textwidth]{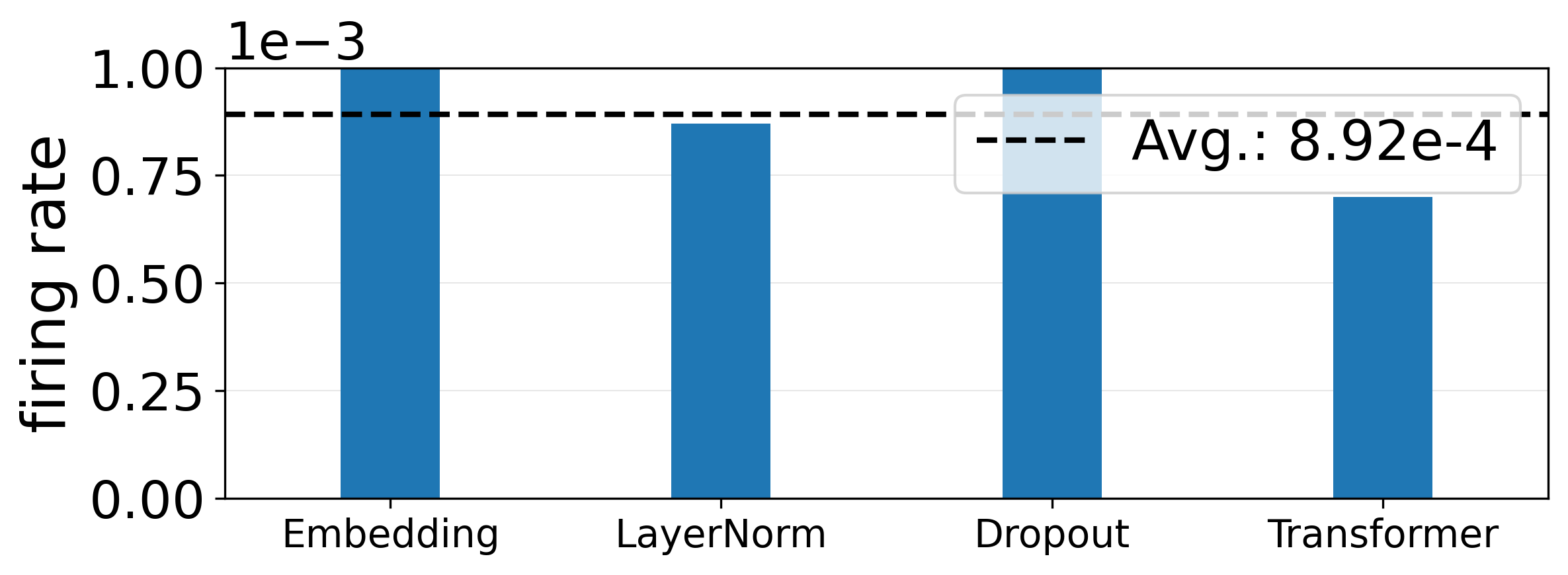}
	\caption{Firing rate in TTFS-BERT.}
	\label{fi}
\end{figure}

We set the time window to 1000 ms and report the firing rate in kHz. We obtain the mean firing rates of the embedding layers, LayerNorm, dropout, and transformer blocks, as shown in ~Figure \ref{fi}, where the firing rate is approximately $8.92 \times 10^{-4}$kHz.
 For reference, rate-based SNN-LLMs report values between $0.1$ and $0.6$ \cite{su2024snn, bal2024spikingbert, xing2024spikelm}, although the original papers do not specify units.
\setlength{\abovedisplayskip}{0.1in}
\setlength{\belowdisplayskip}{0.1in}

\end{document}